\documentclass[10pt,twocolumn,letterpaper]{article}

\usepackage{cvpr}              % To produce the CAMERA-READY version
\usepackage{microtype}
\usepackage[sort&compress,numbers]{natbib}
\usepackage{mathrsfs}
\usepackage{amsmath}
\usepackage{bm}
\usepackage{multirow}
\usepackage{adjustbox}
\usepackage{graphicx}
\usepackage{colortbl}
\usepackage{arydshln}
\usepackage{microtype} % not to line feed as long as it can
\usepackage{amsmath, amssymb, mathtools}
\usepackage{appendix}
\usepackage{multirow}
\usepackage{wrapfig}
\usepackage{multirow}
\definecolor{mygray}{gray}{0.6}
\definecolor{cvprblue}{rgb}{0.21,0.49,0.74}
\definecolor{cvprblue}{rgb}{0.21,0.49,0.74}
\usepackage[pagebackref,breaklinks,colorlinks,allcolors=cvprblue]{hyperref}

\def\paperID{10480} % *** Enter the Paper ID here
\def\confName{CVPR}
\def\confYear{2025}

\title{Targeted Forgetting of Image Subgroups in CLIP Models}

\author{
Zeliang Zhang$^1$ \quad Gaowen Liu$^2$ \quad  Charles Fleming$^2$ \quad Ramana Rao  Kompella$^2$ \quad Chenliang Xu$^1$\\
$^1$University of Rochester \quad $^2$Cisco Research\\
{\tt\small \{zeliang.zhang, chenliang.xu\}@rochester.edu, \{gaoliu, chflemin, rkompell\}@cisco.com}
}

\begin{document}
\maketitle

\begin{abstract}
    Foundation models (FMs) such as CLIP have demonstrated impressive zero-shot performance across various tasks by leveraging large-scale, unsupervised pre-training. However, they often inherit harmful or unwanted knowledge from noisy internet-sourced datasets, compromising their reliability in real-world applications. Existing model unlearning methods either rely on access to pre-trained datasets or focus on coarse-grained unlearning (e.g., entire classes), leaving a critical gap for fine-grained unlearning. In this paper, we address the challenging scenario of selectively forgetting specific portions of knowledge within a class—without access to pre-trained data—while preserving the model’s overall performance. We propose a novel three-stage approach that progressively unlearns targeted knowledge while mitigating over-forgetting. It consists of (1) a forgetting stage to fine-tune the CLIP on samples to be forgotten, (2) a reminding stage to restore performance on retained samples, and (3) a restoring stage to recover zero-shot capabilities using model souping. Additionally, we introduce knowledge distillation to handle the distribution disparity between forgetting/retaining samples and unseen pre-trained data. Extensive experiments on CIFAR-10, ImageNet-1K, and style datasets demonstrate that our approach effectively unlearns specific subgroups while maintaining strong zero-shot performance on semantically similar subgroups and other categories, significantly outperforming baseline unlearning methods, which lose effectiveness under the CLIP unlearning setting. 
    % The project page is \url{https://zhangaipi.github.io/forget_clip/}. 
\end{abstract}

\section{Introduction}
Foundation models (FMs) are deep neural networks pre-trained on large-scale datasets~\citep{subramanian2024towards}. By employing unsupervised training, these models capture complex patterns and relationships within the data, enabling a deep understanding of various domains~\citep{bengio2012unsupervised}. For example, CLIP~\citep{radford2021learning}, a vision-language model, learns correspondences between natural language and images. Trained on the large-scale LAION-5B dataset~\citep{schuhmann2022laion}, CLIP demonstrates strong zero-shot performance~\citep{guo2023calip,sain2023clip,cheng2021data}. However, its reliability in real-world applications is questionable~\citep{zhang2024can,zhang2024discover}. The training data, sourced from the internet, includes uncensored content such as discriminatory and violent imagery, copyright violations, and personal information~\citep{yang2020towards,guo2024domain,quang2021does}. This problematic content can be unintentionally learned during pre-training, raising concerns about the reliability and safety of downstream models~\citep{chen2024catastrophic,li2023whac}. Therefore, it is critical to remove such harmful knowledge to ensure trustworthy FMs.

\begin{figure}
    \centering
    \includegraphics[width=\linewidth]{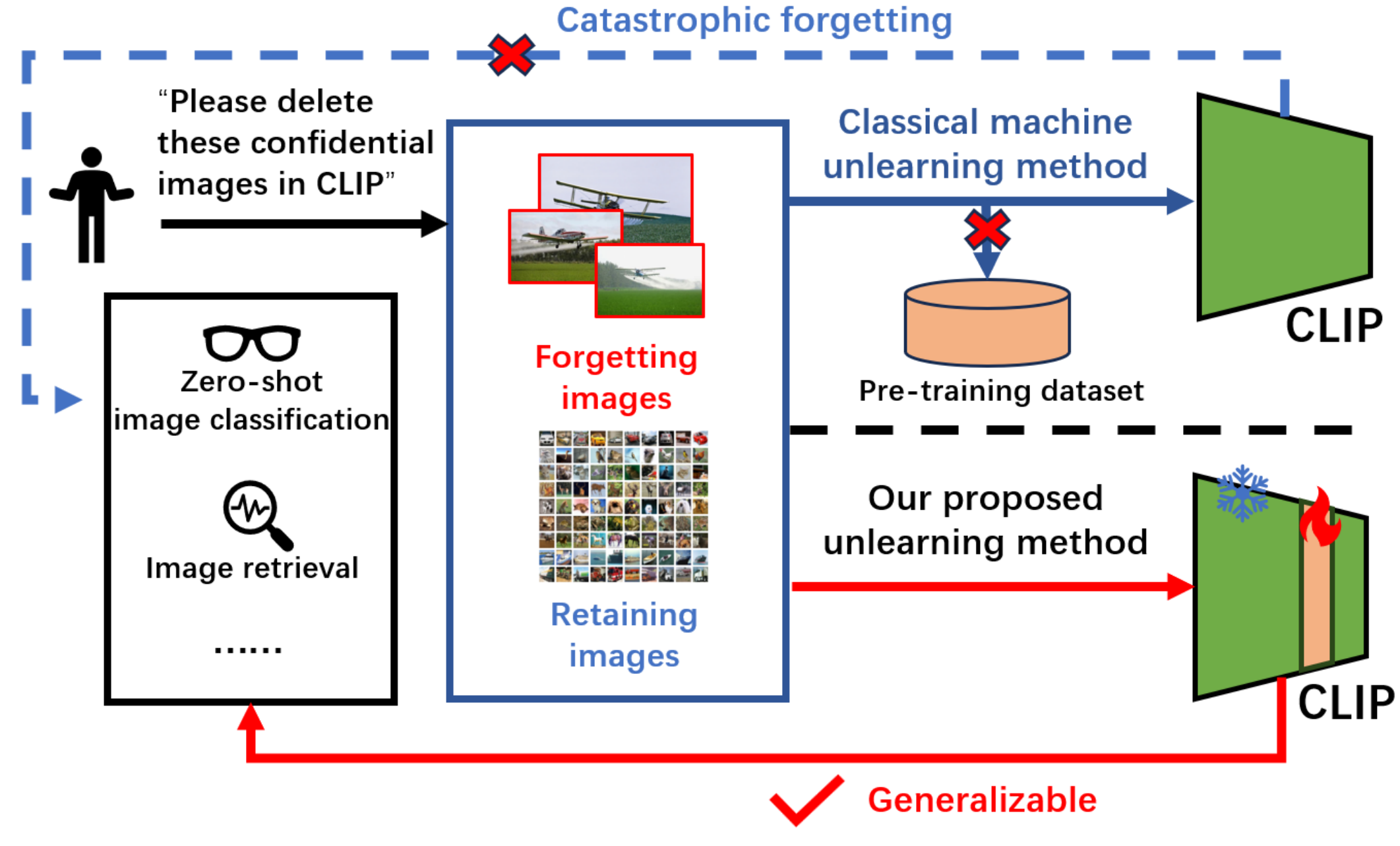}
    \caption{In the task of CLIP unlearning, classical machine unlearning methods often fail to generalize across multiple datasets due to limited access to the original pre-training data. In contrast, our proposed method does not require access to the pre-training dataset. After unlearning, CLIP not only forgets the specified subgroup of images but also maintains consistent performance on various unseen datasets in the zero-shot classification setting.}
    \label{fig:teaser}
\end{figure}

Numerous studies have explored model unlearning, which can be broadly divided into exact~\citep{schelter2023forget,yan2022arcane,ullah2021machine} and approximate strategies~\citep{nguyen2020variational,chien2022efficient,gupta2021adaptive}. Exact unlearning aims to completely eliminate the influence of forgotten data from the model, often by retraining it~\citep{yan2022arcane,zhang2024graph,bourtoule2021machine}, ensuring the model behaves as if the data had never been included in training. While thorough, this approach is computationally expensive and impractical for large-scale models~\citep{liu2024model,sekhari2021remember}. In contrast, approximate unlearning offers a more efficient alternative, using techniques like variational inference~\citep{nguyen2020variational,gong2021bayesian} or adaptive regularization~\citep{zhang2024towards,huang2024unified,mehta2022deep} to partially reduce the data's impact. However, these methods face several challenges when applied to FMs.

\underline{First}, as shown in \cref{fig:teaser}, the training data for FMs is often vast or inaccessible~\citep{schuhmann2022laion,birhane2024into}, rendering exact unlearning impractical. \underline{Second}, while FMs are trained on large-scale datasets using unsupervised learning, the fine-tuning typically used in approximate unlearning strategies~\citep{huang2024learning,tarun2023fast} can degrade generalization across other knowledge domains~\citep{poth2021pre,bai2024pretrain,liu2022improved}, leading to significant performance drops, particularly in zero-shot tasks~\citep{wei2023improving}. \underline{Third}, the knowledge to be forgotten is often deeply entangled with other learned information~\citep{eldan2023s,shostack2024boy}. This entanglement makes it difficult to isolate and remove specific knowledge from the model parameters, particularly when distributional disparities exist between the data we want to unlearn and the pre-trained data.

Few studies have specifically addressed the challenge of unlearning in foundational models, particularly in CLIP. Most existing approaches focus on unlearning entire classes or broad concepts, yet real-world applications often require a more targeted approach, where only certain portions of knowledge within a class are forgotten. For example, we may want the model to forget representations of Disney's Mickey Mouse while retaining versions created by the public, a task known as subgroup image forgetting. Moreover, many current unlearning methods rely on access to the original pre-trained dataset to restore model performance, which is often unavailable. This limitation underscores a critical question in current research: how can models like CLIP achieve selective, fine-grained unlearning without relying on pre-trained data, while still maintaining generalization across other tasks?

In this work, we bridge this gap by tackling a more challenging and practical scenario: enabling CLIP to selectively forget specific portions of knowledge within a single class without access to pre-trained data, framing this as a fine-grained unlearning task. Rather than unlearning an entire class or concept (e.g., all images of airplanes), we focus on selectively forgetting images from a particular subset (e.g., Boeing airplanes) while preserving the model’s knowledge of other subsets (e.g., different manufacturers). This selective forgetting is particularly challenging due to the feature entanglement across subsets within a class, and it is exceedingly difficult to define and collect all subsets within the same class in a foundational model setting, which is trained on vast, large-scale data.

% Unlike unlearning an entire class, where representations are more distinct, fine-grained unlearning requires the model to disentangle overlapping features without degrading its broader understanding of the class. The challenge lies in isolating and removing only the targeted knowledge while maintaining the rest of the model's performance.
%  Both forgetting and retaining samples are drawn from the same distribution.

To address the challenge of partial knowledge removal, we propose a novel, three-stage approach: forgetting, reminding, and restoring. In the forgetting stage, we identify layers essential for representing the forgetting samples but less critical to other data, fine-tuning the model on these samples to selectively remove targeted knowledge. We observe that models can over-forget samples from similar distributions, so we introduce a retaining set with a similar distribution to the targeted subgroup, aiding layer identification and supporting the reminding stage. During reminding, we fine-tune the model on the retaining samples to reinforce CLIP's performance on data prone to over-forgetting due to distributional similarity. In the final restore stage, we employ model souping to fully recover CLIP's zero-shot performance. Throughout this process, we address the distributional gap between forgetting/retaining samples and unseen pre-trained data by distilling knowledge from CLIP, achieving a more precise representation without relying on the original pre-trained dataset. Extensive experiments demonstrate our method’s effectiveness in selectively unlearning subgroup images while preserving robust zero-shot performance across diverse distributions and datasets.

We summarize our contributions as follows: \begin{enumerate} 
\item We formulate the problem of subgroup image unlearning within the CLIP model. 
\item We propose a three-stage method for unlearning in CLIP, consisting of forgetting, reminding, and restoring. Our method is data-efficient, which does not require access to CLIP's pre-training dataset, making it more practical for real-world applications. 
\item We conduct extensive experiments to validate the effectiveness of our proposed method, including unlearning tasks on CIFAR-10, ImageNet-1K, and style forgetting. Additionally, we demonstrate the scalability of our approach across various applications through retrieval-based evaluations. 
\end{enumerate}

\section{Related work}
\noindent \textbf{Machine Unlearning}. Various strategies have been developed to facilitate machine unlearning in models. Notably, \citet{golatkar2020eternal,Alexander23ft} employ a method of continuous fine-tuning (FT) on the retain dataset—those data elements the model is expected to remember. This approach not only approximates the forgetting of the target dataset but also mitigates the issue of overforgetting~\citep{liu2024model,parisi2019continual}, in contrast to exclusively fine-tuning on the forgetting dataset. \citet{becker2022evaluating,liu2024model} leverage Fisher information, which assesses the sensitivity of parameters concerning the forgetting dataset, to achieve model unlearning. Meanwhile, \citet{thudi2022unrolling} unroll the knowledge embedded in the forgetting dataset from the model parameters using stochastic gradient ascent (GA). Further, \citep{foster2024zero} introduces the use of local Lipschitz (LIP) perturbations on the forgetting dataset as a technique for unlearning in image classification models, which \citet{kravets2024zero} subsequently adapts to the CLIP model. Additionally, \citet{chundawat2023zero} proposes an error minimization-maximization (EMMN) framework that aims to maximize errors on the forgetting dataset while minimizing them on the retraining dataset, thus achieving a balance between forgetting and retaining performance. Among these methods, EMMN~\citep{chundawat2023zero} and LIP~\citep{kravets2024zero} stand out as they do not require access to the training datasets (both forgetting and retaining), unlike other approaches.

\noindent \textbf{On the Importance of Unlearning in CLIP}. Trained on a billion-scale dataset of image-text pairs~\citep{schuhmann2022laion}, the CLIP model~\citep{radford2021learning} excels in zero-shot classification tasks. Despite its strengths, the presence of uncensored samples within the training data has raised practical concerns~\citep{msn_article_2024}. Given the impracticality of retraining the CLIP model from scratch due to significant resource demands and the inaccessibility of the large-scale dataset, it is imperative to explore machine unlearning methods to selectively erase problematic knowledge from the model. \citet{poppi2024safe} suggest an approach to first distinguish between safe and unsafe content, followed by fine-tuning CLIP to alter the embeddings of unsafe content. Meanwhile, \citet{cai2024single} focuses on identifying and unlearning from the most influential layer in CLIP, using a subset of the training set to restore zero-shot capabilities. \citep{kravets2024zero} expand zero-shot machine unlearning techniques to the CLIP model, enabling the forgetting of entire concepts as indicated by prompts. Concurrently, \citet{kravets2024zero} apply Layer-wise Relevance Propagation (LoRA) to the text encoder of CLIP, specifically fine-tuning this component to unlearn a given concept within the model.

\section{Problem Formulation}
The CLIP model, denoted \( g(\cdot, \cdot) = \{g^{\text{img}}(\cdot), g^{\text{txt}}(\cdot)\} \), processes image and text inputs into embeddings, \( e^{\text{img}} \) and \( e^{\text{txt}} \). It aims for high cosine similarity between these embeddings when they are correctly paired, and low similarity otherwise, using a threshold \( \delta \).

Suppose a large dataset $D=\{x^\text{img}_{n},x^\text{txt}_{n},c_n,g_n\}_{n=1}^{N}$, where we denote $x^{\text{img}}$ as the image, $x^{txt}$ as the paired text learned by the CLIP, $c$ as the coarse main category, and $g$ is the fine-grained subclass label. The forgetting dataset \( D^f=\{x^{\text{img}}_{n}, x^{\text{txt}}_{n},c_{n},g_{n}\}_{n=1}^{N_f} \) is usually a small set compared with $D$. The goal is to unlearn the associations between $x^{\text{img}}$ and $x^{txt}$ on $D^f$ without impacting the model's performance on the remaining dataset $D^r=D/D_{f}$, namely the retain set. The task involves selectively erasing knowledge from \( D^f \) while maintaining performance on \( D^r \), which can be formulated as follows:
\begin{equation}
\begin{aligned}
    \forall (x^{\text{img}}_{i}, x^{\text{txt}}) \in D^f, \frac{\hat{g}^{\text{img}}(x^{\text{img}}_{i}) \cdot \hat{g}^{\text{txt}}(x^{\text{txt}})}{\|\hat{g}^{\text{img}}(x^{\text{img}}_{i}) \cdot \hat{g}^{\text{txt}}(x^{\text{txt}})\|} < \delta, \\
    \forall (x^{\text{img}}_{i}, x^{\text{txt}}_{i}) \in D^r, \hat{g}(x^{\text{img}}_{i}, x^{\text{txt}}_{i}) \approx g(x^{\text{img}}_{i}, x^{\text{txt}}_{i}),
\end{aligned}\label{eq:task_def}
\end{equation}
where \( \hat{g} \) represents the unlearned CLIP model and \( g \) the original model.

Implementing the approach outlined in \cref{eq:task_def} is challenging due to several factors:
\begin{enumerate}
    \item The large dataset $D$ used for training $g$ is inaccessible, and similarly, $D_r$ is also unavailable. We can only access a small subset of $D_r$ to recover model performance after forgetting, blocking the direct use of conventional machine unlearning methods.
    \item In addition, we may only have access to a coarse label $c$ as the surrogate of $x^{txt}$ rather than a fine-grained subgroup label $g$. This limitation increases the risk of inaccurately erasing targeted knowledge and over-forgetting unrelated knowledge that is similar to the target.
    \item There may be a substantial distribution shift between $D_f$ and $D$. With a large number of parameters, directly fine-tuning on the small dataset $D_f$ could cause the model to overfit and lose generalization ability.
\end{enumerate}

% While procuring images designated for forgetting is relatively straightforward, accurately pairing these images with corresponding texts is challenging due to the lack of access to the complete training dataset. This often compels a reliance on coarse labels, which may inadvertently include images that should not be unlearned. Consequently, this necessitates a shift from whole-concept unlearning to more refined subgroup image unlearning strategies. Furthermore, the task of identifying a suitable retain set that can effectively restore model performance is complicated by the biased and limited nature of the datasets that are typically available.

Thus, the goal of this task is twofold: it involves selectively forgetting only specific image subgroups because of coarse labeling, while simultaneously preserving the model’s robust zero-shot capabilities across various datasets, including those characterized by similarly coarse labels.

\section{Methodology}
\begin{figure*}
    \centering
    \includegraphics[width=0.8\linewidth]{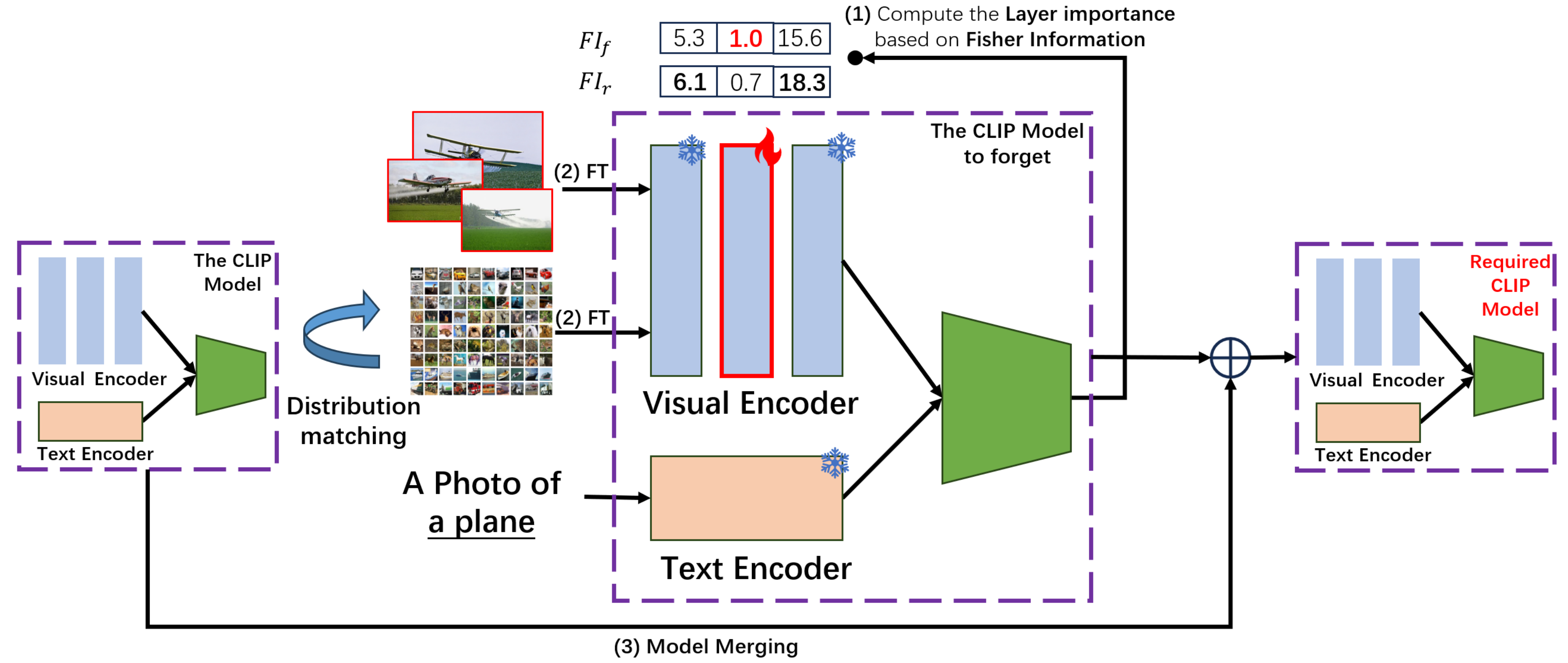}
    \caption{Overview of our method to unlearn the CLIP on a subgroup of images. We first compute the relative fisher information on the forgetting dataset and retain dataset to evaluate the layer importance. We select important layers for the following fine-tuning process. During the fine-tuning process for knowledge reminding, we align the distribution of our retain dataset to that of the pre-training dataset learned by the model. A model merging operation is performed at the last step of knowledge restoration.  }
    \label{fig:method_overview}
\end{figure*}

To address the challenge of subgroup image unlearning, we propose a novel methodology that comprises three stages: forgetting, reminding, and restoring, as depicted in \cref{fig:method_overview}.

\subsection{Forgetting }
Previous studies~\citep{clavera2024machine,liu2023unlearning} indicate that selectively fine-tuning critical layers for a target dataset can effectively mitigate over-forgetting~\citep{liu2024fisher}, typically using Fisher information~\citep{ly2017tutorial,yamashitaone} to assess the importance of each layer. However, while Fisher information measures parameter sensitivity to the target dataset, it overlooks potential impacts on other datasets, which are inaccessible in our task. Our experiments reveal that relying solely on the forgetting dataset to modify the CLIP model often leads to catastrophic over-forgetting on similar subgroups with distributions akin to the forgetting data, as shown in \cref{tab:breeds_forget} for the CLIP model optimized by GA. To address this, we propose using relative Fisher information which combines both the forgetting and retaining samples as a more nuanced metric for identifying essential layers for fine-tuning, thereby enhancing the precision of the unlearning process.

Concretely, given the forgetting dataset $ \mathcal{D}^f = \{(x^{\text{img}}_i, x^{\text{txt}})\} $, we manually construct a retain dataset $ \mathcal{D}^{r} = \{(x^{\text{img}}_i, x^{\text{txt}}_i)\} $ with a distribution similar to $ \mathcal{D}^{f} $, used for alleviating the aforementioned subgroup over-forgetting. Here, $ \mathcal{D}^{r} $ is comprised of similar subgroups of images that we aim to preserve in the model’s memory. In practical scenarios, we often use coarse labels as surrogate image descriptions, so $ x^{\text{txt}} $ from the forgetting dataset may also appear in $ \mathcal{D}^{r} $. We calculate the relative Fisher information for the $ l $-th layer in $ g^{\text{img}}(\cdot) $ as follows:
\begin{equation}
    \mathcal{I}^{l} = \frac{\mathbb{E}_{(x^{\text{img}}_i, x^{\text{txt}}) \sim \mathcal{D}^{f}} \left[ \nabla^2_{\theta^l} (g^{\text{img}}(x^{\text{img}}_i) \cdot g^{\text{txt}}(x^{\text{txt}})) \right]}{\mathbb{E}_{(x^{\text{img}}_i, x^{\text{txt}}_i) \sim \mathcal{D}^{r}} \left[ \nabla^2_{\theta^l} (g^{\text{img}}(x^{\text{img}}_i) \cdot g^{\text{txt}}(x^{\text{txt}}_i)) \right]},
    \label{eq:relative_fisher}
\end{equation}
where $ \theta^l $ represents the parameters of the $ l $-th layer. 

% This approach is tailored for subgroup image unlearning, where coarse labels mean that text representations may be shared across forgetting and retaining datasets, prompting us to focus on unlearning within the image encoder of CLIP.

We then rank the importance of different layers based on the computed relative Fisher information, as detailed in \cref{eq:relative_fisher}. Layers with higher relative importance are selected for targeted unlearning, with all other layers frozen to minimize the impact on non-target knowledge. In our experiments, applying the low-rank adaptation (LoRA) technique~\citep{hu2021lora} to the selected layers significantly enhances the targeted unlearning performance while minimizing effects on other retained knowledge.

Finally, we fine-tune our model on the forgetting dataset $\mathcal{D}_{f}$ to minimize the following loss function:
\begin{equation}
    \min \mathcal{L}_{f} = \sum_{i} \frac{g^{img}(x^{img}_{i})\cdot g^{txt}(x^{txt})}{\Vert g^{img}(x^{img}_{i})\cdot g^{txt}(x^{txt})\Vert},
    \label{eq:forget}
\end{equation}
where images in the forgetting dataset are associated through the same coarse label $x^{txt}$.

By fine-tuning the CLIP model on images marked for unlearning with coarse labels, we aim to selectively erase specific knowledge. However, this adjustment often leads to catastrophic overforgetting, affecting the model's ability across both similar and different image distributions. For example, attempting to forget the "airplane" class in CIFAR-10 may also affect the recognition of other classes within CIFAR-10 and in ImageNet-1K. To tackle these challenges, particularly when full access to the pre-training dataset is unavailable, we implement reminding and restoration stages.

\subsection{Reminding}
An intuitive strategy to retain non-targeted knowledge in CLIP is to fine-tune it on our retaining set, built during the forgetting stage for computing $\mathcal{I}$. However, two challenges hinder the direct application of this fine-tuning step. First, our retaining set may have a different distribution than the pre-training dataset, potentially causing inefficient learning. Second, our retaining set is typically limited, and fine-tuning on it could lead to further catastrophic over-forgetting in various unseen datasets and tasks.

To address the first challenge, inspired by research showing that training data features are encoded in model parameters~\citep{yin2024squeeze}, we employ an optimization-based strategy to align the retaining dataset’s distribution with the pre-training dataset. Specifically, for a given batch of image-text pairs in the retaining dataset $ \mathcal{D}_r = \{(x^{\text{img}}_{i}, x^{\text{txt}}_{i})\}_{i=1}^{B} $, we introduce a batch of perturbations $ {\delta_i}_{i=1}^{B} $ on the images. These perturbations are optimized by minimizing the loss:
\begin{equation}
    \begin{aligned}
        \mathcal{L}_a = &\sum_{l} \left\| \frac{1}{B} \sum_{i=1}^{B} \mu^{img}_{l}(x^{img}_{i} + \delta_{i}) - \text{BN}^{\mu}_{l} \right\| \\
        +& \sum_{l} \left\| \frac{1}{B} \sum_{i=1}^{B} \sigma^{img}_{l}(x^{img}_{i} + \delta_i) - \text{BN}^{\sigma}_{l} \right\|,
    \end{aligned}
    \label{eq:distill}
\end{equation}
where $ \mu^{img}_{l}(\cdot) $ and $ \sigma^{img}_{l} $ are the mean and variance of the batch of intermediate features before the $ l $-th BatchNormalization (BN) layer, computed by the image encoder of original CLIP. $ \text{BN}^{\mu}_{l} $ and $ \text{BN}^{\sigma}_{l} $ are the mean and variance parameters of the $ l $-th BN layer, encoding the global distribution information of the pre-training dataset.

We then fine-tune the CLIP model on the retaining dataset, now aligned with the pre-training dataset’s distribution. This serves as a reminder of the knowledge encoded in the original model, effectively achieved without direct access to the pre-training dataset. This strategy helps prevent further catastrophic forgetting that could arise from distributional disparities between the retaining dataset and the original training dataset.

To address the second challenge of overfitting during fine-tuning due to the large amount of parameters, we propose minimizing the impact on the original model parameters by using an exponential moving average (EMA)~\citep{klinker2011exponential} for the final model parameters, as follows:

\begin{equation} \begin{aligned} \theta^{\text{ema}} = \alpha \cdot \theta^{\text{ema}} + (1 - \alpha) \cdot \theta, \end{aligned} \label{eq} \end{equation}
where $ \theta^{\text{ema}} $ is initialized as the original CLIP parameters $ \theta_{0} $, $ \theta $ represents the continuously optimized parameters from $ \theta_{0} $, and $ \alpha $ is the decay factor. This approach effectively balances parameter retention with necessary updates.

% Because it is hard to access the pre-training dataset, it is essential to consider fully exploiting the retaining dataset at hand for reminding.  We propose the approach of distribution matching to transform the images in the retaining dataset  towards the optima approaching distribution of the knowledge learned by the CLIP at the pre-training process.  

\subsection{Restoration}
In addition to fine-tuning on the retain dataset for reminding, we also employ a model merging strategy to restore the zero-shot performance of CLIP, a technique empirically shown to locate flat optima on the loss landscape, enhancing generalization~\citep{wortsman2022model}. 

Specifically, we first compile a small calibration dataset $\mathcal{D}_{m} = \{(x_i, y_i)\}_{i=1}^{N}$ for zero-shot image classification, which may include portions of the retain dataset not used in fine-tuning. We then execute model merging with varying coefficients as follows:
\begin{equation}
    \begin{aligned}
        \arg\max_{\theta}~~~ &\text{Acc}(g, \mathcal{D}_{m}),\\
        \textit{s.t.} ~~ \theta = \alpha &\cdot \theta^{f} + (1-\alpha) \cdot \theta^{ori},
    \end{aligned}\label{eq:merge}
\end{equation}
where $\textit{Acc}(g, \mathcal{D}_{m})$ assesses the zero-shot classification performance of model $g$ on $\mathcal{D}_{m}$, and $\theta^{f}$ and $\theta^{ori}$ represent the parameters of the CLIP model post-unlearning and the original model, respectively.

\section{Evaluations}

% 区分OOD集
% Score用蓝色标注   
% Target 直接标注100-x
% 取消上下箭头
% 注意结果黑体
\begin{table*}[]
\caption{Evaluation results for forgetting the subclass 'marmoset' within the superclass 'monkey' (left) and 'box turtle' within the superclass 'turtle' (right) in the ImageNet-1K dataset are presented (Target). Each superclass contains four subgroups, with the remaining three designated as the retain set (Retain).}\label{tab:breeds_forget}
\begin{subtable}{0.48\textwidth}
\centering
\resizebox{\linewidth}{!}{
\begin{tabular}{c|c|ccccccc|c}
\toprule
\multirow{2}{*}{Backbone} &
  \multirow{2}{*}{Method} &
  \multicolumn{3}{c}{ImageNet} &
  \multirow{2}{*}{CIFAR} &
  \multirow{2}{*}{Food} &
  \multirow{2}{*}{STL} &
  \multirow{2}{*}{ObjectNet}&
   \multirow{2}{*}{Score}\\
                       &               & Target ($\downarrow$)       & Retain ($\uparrow$)  & All ($\uparrow$)       &               &               &                       &      \\\hline
\multirow{7}{*}{RN50}  & \textcolor{mygray}{Original}      & \textcolor{mygray}{51.0}         & \textcolor{mygray}{54.7}          & \textcolor{mygray}{59.8}          & \textcolor{mygray}{70.4}          & \textcolor{mygray}{60.9}          & \textcolor{mygray}{92.0}          & \textcolor{mygray}{68.9}   & --       \\
& FT   & ~2.5$_{\textcolor{blue}{4.9}}$  & 87.1$_{\textcolor{blue}{100.0}}$ & 25.6$_{\textcolor{blue}{42.8}}$ & 13.5$_{\textcolor{blue}{19.2}}$  & 17.4$_{\textcolor{blue}{28.5}}$   & 43.1$_{\textcolor{blue}{46.8}}$   & 19.6$_{\textcolor{blue}{28.4}}$  & 51.6    \\
& GA   & ~0.0$_{\textcolor{blue}{0.0}}$  & ~0.9$_{\textcolor{blue}{1.6}}$ & 32.2$_{\textcolor{blue}{53.8}}$  & 16.4$_{\textcolor{blue}{23.3}}$  & 22.9$_{\textcolor{blue}{37.6}}$   & 63.3$_{\textcolor{blue}{68.8}}$   & 22.1$_{\textcolor{blue}{32.1}}$   & 45.3   \\
& Fisher    &  ~0.2$_{\textcolor{blue}{0.3}}$ & ~0.5$_{\textcolor{blue}{0.9}}$ & ~1.3$_{\textcolor{blue}{2.1}}$  & 10.4$_{\textcolor{blue}{14.7}}$  & ~0.2$_{\textcolor{blue}{0.3}}$   & 10.5$_{\textcolor{blue}{11.4}}$   & ~3.0$_{\textcolor{blue}{4.4}}$     & 19.1 \\
& LIP    &  ~0.7$_{\textcolor{blue}{1.4}}$ & ~0.2$_{\textcolor{blue}{3.6}}$ & ~1.3$_{\textcolor{blue}{2.2}}$  & 10.7$_{\textcolor{blue}{15.2}}$  & ~0.2$_{\textcolor{blue}{0.3}}$   & 10.3$_{\textcolor{blue}{11.2}}$   & ~1.6$_{\textcolor{blue}{2.3}}$    &  18.6 \\
& EMMN  &  ~0.0$_{\textcolor{blue}{0.0}}$ & 56.7$_{\textcolor{blue}{100.0}}$ & 28.4$_{\textcolor{blue}{47.5}}$ & 13.1$_{\textcolor{blue}{18.6}}$  & 20.4$_{\textcolor{blue}{33.5}}$   & 54.9$_{\textcolor{blue}{59.7}}$   & 20.5$_{\textcolor{blue}{29.7}}$    & 55.6  \\
& CLIP-LIP  &  2.6$_{\textcolor{blue}{5.1}}$ & 0.4$_{\textcolor{blue}{0.7}}$ & 17.7$_{\textcolor{blue}{29.6}}$ & 18.8$_{\textcolor{blue}{26.7}}$  &  4.2$_{\textcolor{blue}{6.9}}$  & 50.8$_{\textcolor{blue}{55.2}}$   &  12.5$_{\textcolor{blue}{18.1}}$  & 33.2   \\
& \textbf{Ours} &  \textbf{~0.0}$_{\textcolor{blue}{0.0}}$ & \textbf{50.2}$_{\textcolor{blue}{91.7}}$ & \textbf{54.5}$_{\textcolor{blue}{91.1}}$  & \textbf{85.7}$_{\textcolor{blue}{100.0}}$ & \textbf{62.9}$_{\textcolor{blue}{100.0}}$ & \textbf{81.5}$_{\textcolor{blue}{88.6}}$ & \textbf{45.2}$_{\textcolor{blue}{65.6}}$  & \textbf{91.0}  \\\hline
\multirow{7}{*}{RN101} & \textcolor{mygray}{Original}      & \textcolor{mygray}{66.2}         & \textcolor{mygray}{52.6}          & \textcolor{mygray}{62.3}          & \textcolor{mygray}{64.4}          & \textcolor{mygray}{57.3}          & \textcolor{mygray}{92.4}          & \textcolor{mygray}{69.2}   & --       \\
& FT    & ~1.3$_{\textcolor{blue}{2.0}}$  & 75.2$_{\textcolor{blue}{100.0}}$ & 27.6$_{\textcolor{blue}{44.3}}$  & 11.8$_{\textcolor{blue}{18.3}}$  & ~4.4$_{\textcolor{blue}{7.6}}$   & 35.1$_{\textcolor{blue}{38.0}}$   & 17.3$_{\textcolor{blue}{25.0}}$   & 47.3   \\
& GA  & ~0.7$_{\textcolor{blue}{1.1}}$  & ~0.3$_{\textcolor{blue}{0.5}}$ & 36.2$_{\textcolor{blue}{58.1}}$  & 19.9$_{\textcolor{blue}{30.9}}$  & 29.3$_{\textcolor{blue}{51.1}}$   & 58.6$_{\textcolor{blue}{63.4}}$   & 19.4$_{\textcolor{blue}{28.0}}$   & 47.3   \\
& Fisher   &  ~0.9$_{\textcolor{blue}{1.4}}$ & ~0.4$_{\textcolor{blue}{0.7}}$ & ~0.3$_{\textcolor{blue}{0.5}}$  & 11.7$_{\textcolor{blue}{18.2}}$  & ~0.2$_{\textcolor{blue}{0.3}}$   & ~9.6$_{\textcolor{blue}{10.3}}$   & ~1.0$_{\textcolor{blue}{1.4}}$   & 18.6   \\
& LIP   &  ~0.2$_{\textcolor{blue}{0.3}}$ & ~0.3$_{\textcolor{blue}{0.5}}$ & ~2.0$_{\textcolor{blue}{3.2}}$  & 10.4$_{\textcolor{blue}{16.1}}$  & ~0.2$_{\textcolor{blue}{0.3}}$   & 10.7$_{\textcolor{blue}{11.6}}$   & ~1.9$_{\textcolor{blue}{2.7}}$   & 19.2   \\
& EMMN   & ~0.7$_{\textcolor{blue}{1.1}}$  & 62.5$_{\textcolor{blue}{100.0}}$ & 30.1$_{\textcolor{blue}{48.3}}$ & 22.6$_{\textcolor{blue}{35.1}}$  & 10.8$_{\textcolor{blue}{18.8}}$   & 10.0$_{\textcolor{blue}{10.8}}$   & 23.8$_{\textcolor{blue}{34.4}}$   &  49.5 \\
& CLIP-LIP  &  2.7$_{\textcolor{blue}{4.1}}$ & 1.5$_{\textcolor{blue}{2.9}}$ & 32.0$_{\textcolor{blue}{51.4}}$ & 28.6$_{\textcolor{blue}{44.4}}$  & 10.9$_{\textcolor{blue}{19.0}}$   & 49.6$_{\textcolor{blue}{53.7}}$   &  14.9$_{\textcolor{blue}{21.5}}$   & 41.2  \\
& \textbf{Ours} &  \textbf{~0.0}$_{\textcolor{blue}{0.0}}$ & \textbf{47.7}$_{\textcolor{blue}{90.7}}$ &  \textbf{58.5}$_{\textcolor{blue}{93.9}}$ & \textbf{69.7}$_{\textcolor{blue}{100.0}}$ & \textbf{56.9}$_{\textcolor{blue}{99.3}}$ & \textbf{93.5}$_{\textcolor{blue}{100.0}}$ & \textbf{45.9}$_{\textcolor{blue}{66.3}}$  & \textbf{92.9}\\\bottomrule
\end{tabular}}
\end{subtable}
\begin{subtable}{0.48\textwidth}
\centering
\resizebox{\linewidth}{!}{
\begin{tabular}{c|c|ccccccc|c}
\toprule
\multirow{2}{*}{Backbone} &
  \multirow{2}{*}{Method} &
  \multicolumn{3}{c}{ImageNet} &
  \multirow{2}{*}{CIFAR} &
  \multirow{2}{*}{Food} &
  \multirow{2}{*}{STL} &
  \multirow{2}{*}{ObjectNet} & \multirow{2}{*}{Score} \\
                       &               & Target ($\downarrow$)        & Retain ($\uparrow$)  & All ($\uparrow$)       &               &               &                    &        \\\hline
\multirow{7}{*}{RN50}   & \textcolor{mygray}{Original}      & \textcolor{mygray}{64.5}         & \textcolor{mygray}{70.3}          & \textcolor{mygray}{59.8}          & \textcolor{mygray}{70.4}          & \textcolor{mygray}{60.9}          & \textcolor{mygray}{92.0}          & \textcolor{mygray}{68.9}    & --      \\
& FT   &  ~0.0$_{\textcolor{blue}{0.0}}$ & 72.1$_{\textcolor{blue}{100.0}}$ & 26.0$_{\textcolor{blue}{43.5}}$  & 12.6$_{\textcolor{blue}{17.9}}$  & 28.6$_{\textcolor{blue}{47.0}}$   & 70.3$_{\textcolor{blue}{76.4}}$   & 20.2$_{\textcolor{blue}{29.3}}$    & 59.2  \\
& GA   & ~0.0$_{\textcolor{blue}{0.0}}$  & 13.3$_{\textcolor{blue}{18.9}}$ & 42.1$_{\textcolor{blue}{70.4}}$  & 14.6$_{\textcolor{blue}{20.7}}$  & 40.0$_{\textcolor{blue}{65.7}}$   & 77.8$_{\textcolor{blue}{84.6}}$   & 21.3$_{\textcolor{blue}{30.9}}$    & 55.8  \\
& Fisher  &  ~1.2$_{\textcolor{blue}{1.9}}$ & ~2.0$_{\textcolor{blue}{2.8}}$  & ~3.5$_{\textcolor{blue}{5.9}}$   & ~9.7$_{\textcolor{blue}{13.8}}$  & ~0.4$_{\textcolor{blue}{0.7}}$   & 10.5$_{\textcolor{blue}{11.4}}$   & ~1.6$_{\textcolor{blue}{2.3}}$    & 19.3  \\
& LIP  & ~0.4$_{\textcolor{blue}{0.6}}$  & ~3.4$_{\textcolor{blue}{4.8}}$ & 15.2$_{\textcolor{blue}{25.4}}$   & 10.6$_{\textcolor{blue}{15.1}}$  & ~6.1$_{\textcolor{blue}{10.0}}$   & 23.7$_{\textcolor{blue}{25.8}}$   & 11.3$_{\textcolor{blue}{16.4}}$   & 28.1   \\
& EMMN  &  ~0.1$_{\textcolor{blue}{0.2}}$ & 57.2$_{\textcolor{blue}{81.4}}$ & 28.1$_{\textcolor{blue}{47.0}}$ & 12.5$_{\textcolor{blue}{17.8}}$  & 22.1$_{\textcolor{blue}{36.3}}$   & 64.7$_{\textcolor{blue}{70.3}}$   & 17.6$_{\textcolor{blue}{25.5}}$   & 54.0   \\
& CLIP-LIP  &  1.9$_{\textcolor{blue}{2.9}}$ & 2.0$_{\textcolor{blue}{2.8}}$ & 31.9$_{\textcolor{blue}{53.3}}$ & 19.5$_{\textcolor{blue}{27.7}}$  & 6.9$_{\textcolor{blue}{11.3}}$   & 63.6$_{\textcolor{blue}{69.1}}$   &  34.2$_{\textcolor{blue}{49.6}}$   & 44.4  \\
& \textbf{Ours}&  \textbf{~0.0}$_{\textcolor{blue}{0.0}}$ & \textbf{69.4}$_{\textcolor{blue}{98.7}}$ & \textbf{50.6}$_{\textcolor{blue}{84.6}}$  & \textbf{54.5}$_{\textcolor{blue}{77.4}}$ & \textbf{50.5}$_{\textcolor{blue}{82.9}}$ & \textbf{87.6}$_{\textcolor{blue}{95.2}}$ & \textbf{43.1}$_{\textcolor{blue}{62.6}}$ & \textbf{85.9}  \\\hline
\multirow{7}{*}{RN101} & \textcolor{mygray}{Original}      & \textcolor{mygray}{66.2}         & \textcolor{mygray}{77.1}          & \textcolor{mygray}{62.3}          & \textcolor{mygray}{64.4}          & \textcolor{mygray}{57.3}          & \textcolor{mygray}{92.4}          & \textcolor{mygray}{69.2}  & --        \\
& FT    & ~0.0$_{\textcolor{blue}{0.0}}$  &68.0$_{\textcolor{blue}{88.2}}$  & 32.4$_{\textcolor{blue}{52.0}}$ & 11.4$_{\textcolor{blue}{17.7}}$  & 27.8$_{\textcolor{blue}{48.5}}$   & 66.0$_{\textcolor{blue}{71.4}}$   & 23.5$_{\textcolor{blue}{34.0}}$   & 58.8   \\
& GA  & ~0.0$_{\textcolor{blue}{0.0}}$  & 22.1$_{\textcolor{blue}{28.7}}$ & 46.4$_{\textcolor{blue}{74.5}}$ & 12.2$_{\textcolor{blue}{18.9}}$  & 40.4$_{\textcolor{blue}{70.5}}$   & 75.6$_{\textcolor{blue}{81.8}}$   & 25.9$_{\textcolor{blue}{37.4}}$    & 58.8  \\
& Fisher  & ~1.9$_{\textcolor{blue}{2.9}}$  & ~2.4$_{\textcolor{blue}{3.1}}$ & ~0.9$_{\textcolor{blue}{1.4}}$ & 11.0$_{\textcolor{blue}{17.1}}$  & ~0.3$_{\textcolor{blue}{0.5}}$   & ~9.5$_{\textcolor{blue}{10.3}}$   & ~0.6$_{\textcolor{blue}{0.9}}$   & 18.6   \\
& LIP   & ~0.8$_{\textcolor{blue}{1.2}}$  & ~3.2$_{\textcolor{blue}{4.2}}$ & 12.9$_{\textcolor{blue}{20.7}}$  & 10.3$_{\textcolor{blue}{16.0}}$  & ~8.4$_{\textcolor{blue}{14.7}}$   & 16.9$_{\textcolor{blue}{18.3}}$   & 14.7$_{\textcolor{blue}{21.2}}$   & 27.7   \\
& EMMN   &  ~0.0$_{\textcolor{blue}{0.0}}$ & 48.6$_{\textcolor{blue}{63.0}}$ &  32.5$_{\textcolor{blue}{52.2}}$ & ~9.2$_{\textcolor{blue}{14.3}}$  & 22.4$_{\textcolor{blue}{39.1}}$   & 60.8$_{\textcolor{blue}{65.8}}$   & 22.9$_{\textcolor{blue}{33.1}}$   & 52.5   \\
& CLIP-LIP  & 4.6$_{\textcolor{blue}{6.9}}$  & 10.8$_{\textcolor{blue}{14.0}}$ & 38.1$_{\textcolor{blue}{61.2}}$ & 38.5$_{\textcolor{blue}{59.8}}$   & 17.6$_{\textcolor{blue}{30.7}}$   & 66.8$_{\textcolor{blue}{72.3}}$   &  18.6$_{\textcolor{blue}{26.9}}$   & 51.1  \\
& \textbf{Ours}& \textbf{~0.0}$_{\textcolor{blue}{0.0}}$  & \textbf{80.1}$_{\textcolor{blue}{100.0}}$ & \textbf{59.3}$_{\textcolor{blue}{95.2}}$  & \textbf{70.9}$_{\textcolor{blue}{100.0}}$ & \textbf{55.8}$_{\textcolor{blue}{97.4}}$ & \textbf{91.5}$_{\textcolor{blue}{99.0}}$ & \textbf{46.2}$_{\textcolor{blue}{66.8}}$ & \textbf{94.1} \\\bottomrule
\end{tabular}}
\end{subtable}

\end{table*}
\begin{figure*}
    \centering
    \includegraphics[width=0.85\linewidth]{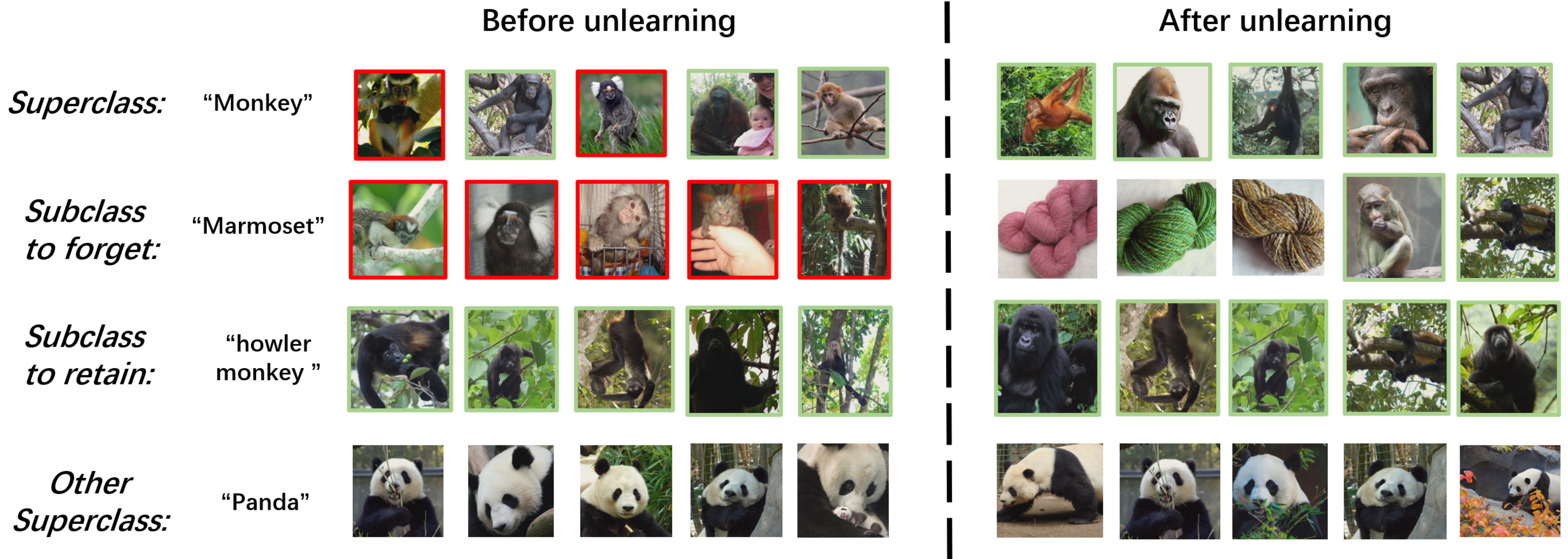}
    \caption{Retrieval results on the ImageNet-1K dataset before and after the unlearning process in the CLIP model. We highlight the "marmoset" subgroup in \textcolor{red}{red}, with other "monkey" subgroups shown in \textcolor{green}{green}.}
    \label{fig:unlearn_monkey}
\end{figure*}
We begin by focusing on the forgetting subgroups within the ImageNet dataset, where the images are semantically grouped by Breeds. We select two superclasses: ‘monkey’ and ‘turtle.’ Each superclass contains four distinct subgroups. We specifically target one subgroup in each superclass for unlearning: ‘marmoset’ in the monkey superclass and ‘box turtle’ in the turtle superclass.

\subsection{Experimental setup} 
\noindent \textbf{Settings}. We conduct experiments in three distinct settings: in-domain forgetting, near-domain forgetting, and style forgetting, as detailed below:
\begin{itemize} 
\item \textit{In-domain forgetting}: Experiments are conducted on the Breeds dataset~\citep{santurkar2020breeds}, a subset of the ImageNet-1K dataset~\citep{deng2009imagenet} consisting of 4 subgroups within 17 superclasses. The images, with dimensions of $224 \times 224$, closely match the distribution of images in the pre-trained LAION-5B dataset used for CLIP. For each superclass, one subgroup is selected for unlearning, while the images from the remaining three subgroups form the retaining set.
\item \textit{Out-of-domain forgetting}: Experiments are performed on the CIFAR-10 dataset~\citep{krizhevsky2009learning}, which differs significantly from the image distribution in LAION-5B. To align with CLIP’s input format, we resize the images from $32 \times 32$ to $224 \times 224$. In this setting, we select a class shared between CIFAR-10 and ImageNet-1K. The images of this selected class form the forgetting dataset, while the remaining classes in CIFAR-10 constitute the retaining set.
\item \textit{Style forgetting}: Experiments are conducted on one class in the ImageNet-1K dataset~\citep{deng2009imagenet} transformed by Stable Diffusion~\citep{chung2024style}. The images are randomly transformed into three styles: sketch style, poster art, and black-and-white drawing. For a given style to forget, images in the remaining styles are treated as the retaining set.
\end{itemize}

\noindent \textbf{Baselines and implementations}. We select six methods as our baselines: FT~\citep{Alexander23ft}, GA~\citep{thudi2022unrolling}, Fisher~\citep{liu2024model}, LIP~\citep{foster2024zero}, and EMMN~\citep{chundawat2023zero},and CLIP-LIP~\citep{kravets2024zero}. Except the CLIP-LIP, these methods are developed for conventional models that require a classification head. To address this difference, we use the probability from the zero-shot classification task as a surrogate for the classification logits. In these approaches, the text prompt is provided by the superclass names in both the forgetting and retain sets. For all learning-based methods, recognizing the sensitivity of CLIP parameters during the fine-tuning process, we use the Adam optimizer with a small learning rate of $1\times 10^{-6}$ for up to 2 epochs. For LIP, we set the number of noisy copies to 10. Additionally, the variance scaling factor $\alpha$ in Fisher is set to 0.2. We provide the full implementation in \cref{sec:imp_base}.

\noindent \textbf{Evaluations}. The performance of image subgroup unlearning in CLIP is evaluated along three dimensions. \textit{First}, we must ensure that CLIP successfully forgets the images from the specified subgroup. \textit{Second}, CLIP must continue to perform consistently on the retain set, which includes the other subgroups that are not intended to be forgotten. \textit{Third}, CLIP must maintain strong zero-shot performance across various tasks and datasets. These considerations lead us to evaluate CLIP's performance not only on the targeted forgetting subgroups and unaffected subgroups, but also on various unseen datasets during the forgetting process, including Food~\citep{bossard2014food}, STL~\citep{coates2011analysis}, and ObjectNet~\citep{barbu2019objectnet}. Additionally, we will use the post-unlearning CLIP to retrieve images in order to assess the impact of forgetting on its learned representations. Additionally, we use the classification accuracy ratio between the original and post-unlearned CLIP models to gauge the degree of knowledge forgetting and retention. More specifically, the ratio is computed as $\min(\frac{\text{Acc}_{unlearn}}{\text{Acc}_{ori}},1)$, where we cap the score at 1 to represent full knowledge restoration. This ratio is shown in the \textcolor{blue}{subscript} of each result. We further compute the mean of these ratios across all classes, denoted as the \textbf{Score}, to provide a comprehensive evaluation metric.

\subsection{In-domain subgroup forgetting in ImageNet}

\begin{table*}[]
\caption{Evaluation results for forgetting the “airplane” (left) and "truck" (right) class in CIFAR-10 are presented. To assess the unaffected subgroup performance, we also separately retrieve the “airplane” class from ImageNet-1K (see the column labeled Target in ImageNet). The CIFAR-10 dataset is used exclusively as the retain set (see the column labeled Retain in CIFAR).}\label{tab:forget_cifar}
\begin{subtable}{0.48\textwidth}
\centering
\resizebox{\linewidth}{!}{
\begin{tabular}{c|c|ccccccc|c}
\toprule
\multirow{2}{*}{Backbone} &
  \multirow{2}{*}{Method} &
  \multicolumn{2}{c}{CIFAR} &
  \multirow{2}{*}{Food} &
  \multirow{2}{*}{STL} &
  \multirow{2}{*}{ObjectNet} &
  \multicolumn{2}{c}{ImageNet} &
   \multirow{2}{*}{Score} \\
                       &               & Target  ($\downarrow$)       & Retain  ($\uparrow$)       &               &               &               & Target        & All    &      \\\hline
\multirow{7}{*}{RN50}  & \textcolor{mygray}{Original}      & \textcolor{mygray}{81.3}         & \textcolor{mygray}{70.4}          & \textcolor{mygray}{60.9}          & \textcolor{mygray}{92.0}          & \textcolor{mygray}{68.9}          & \textcolor{mygray}{53.6}          & \textcolor{mygray}{59.8}     & --     \\
                       & FT            & ~0.0$_{\textcolor{blue}{0.0}}$           & 92.6$_{\textcolor{blue}{100.0}}$           & ~1.1$_{\textcolor{blue}{1.8}}$           & 41.7$_{\textcolor{blue}{45.3}}$          & ~0.2$_{\textcolor{blue}{0.3}}$           & ~0.0$_{\textcolor{blue}{0.0}}$          & ~0.3$_{\textcolor{blue}{0.5}}$   & 35.4        \\
                       & GA            & ~0.0$_{\textcolor{blue}{0.0}}$          & 14.2$_{\textcolor{blue}{20.2}}$          & ~1.0$_{\textcolor{blue}{1.6}}$          & 14.1$_{\textcolor{blue}{15.3}}$          & ~0.1$_{\textcolor{blue}{0.1}}$           & ~0.2$_{\textcolor{blue}{0.4}}$           & ~0.1$_{\textcolor{blue}{0.2}}$    & 19.7      \\
                       & Fisher        & ~0.0$_{\textcolor{blue}{0.0}}$           & 13.1$_{\textcolor{blue}{18.6}}$          & ~1.0$_{\textcolor{blue}{1.6}}$           & 15.2$_{\textcolor{blue}{16.5}}$         & ~0.3$_{\textcolor{blue}{0.4}}$          & ~0.0$_{\textcolor{blue}{0.0}}$         & ~0.1$_{\textcolor{blue}{0.2}}$       & 19.6   \\
                       & LIP           & ~0.0$_{\textcolor{blue}{0.0}}$          & 15.7$_{\textcolor{blue}{22.3}}$          & ~1.3$_{\textcolor{blue}{2.1}}$           & 14.9$_{\textcolor{blue}{16.2}}$         & ~0.3$_{\textcolor{blue}{0.4}}$           & ~0.1$_{\textcolor{blue}{0.2}}$         & ~0.2$_{\textcolor{blue}{0.3}}$     & 20.2     \\
                       & EMMN          & ~0.0$_{\textcolor{blue}{0.0}}$          & 80.7$_{\textcolor{blue}{100.0}}$           & ~1.6$_{\textcolor{blue}{2.6}}$           & 41.6$_{\textcolor{blue}{45.2}}$          & ~0.2$_{\textcolor{blue}{0.3}}$           & ~0.0$_{\textcolor{blue}{0.0}}$          & ~0.5$_{\textcolor{blue}{0.8}}$     & 35.6       \\
                       & CLIP-LIP          &   1.1$_{\textcolor{blue}{1.4}}$       &  27.2$_{\textcolor{blue}{38.6}}$          &    7.1$_{\textcolor{blue}{11.7}}$        &    49.9$_{\textcolor{blue}{54.2}}$      &     31.2$_{\textcolor{blue}{45.3}}$       &       34.5$_{\textcolor{blue}{64.4}}$    &    26.8$_{\textcolor{blue}{44.8}}$     & 51.1    \\
                       & \textbf{Ours} & \textbf{~7.6}$_{\textcolor{blue}{9.3}}$ & \textbf{65.1}$_{\textcolor{blue}{92.5}}$ & \textbf{51.9}$_{\textcolor{blue}{85.2}}$ & \textbf{85.7}$_{\textcolor{blue}{93.2}}$ & \textbf{57.9}$_{\textcolor{blue}{84.0}}$ & \textbf{49.8}$_{\textcolor{blue}{92.9}}$ & \textbf{50.1}$_{\textcolor{blue}{83,8}}$ & \textbf{88.9} \\\hline
\multirow{7}{*}{RN101} & \textcolor{mygray}{Original}      & \textcolor{mygray}{69.2}         & \textcolor{mygray}{64.4}          & \textcolor{mygray}{57.3}          & \textcolor{mygray}{92.4}          & \textcolor{mygray}{69.2}          & \textcolor{mygray}{51.9}          & \textcolor{mygray}{62.3} & --         \\
                       & FT            & ~0.0$_{\textcolor{blue}{0.0}}$           & 94.6$_{\textcolor{blue}{100.0}}$          & ~1.1$_{\textcolor{blue}{1.9}}$          & 58.4$_{\textcolor{blue}{63.2}}$          & ~0.4$_{\textcolor{blue}{0.6}}$           & ~0.0$_{\textcolor{blue}{0.0}}$         & ~0.2$_{\textcolor{blue}{0.3}}$     & 38.0     \\
                       & GA            & ~0.0$_{\textcolor{blue}{0.0}}$          & 11.1$_{\textcolor{blue}{17.2}}$        & ~0.9$_{\textcolor{blue}{1.6}}$           & 15.0$_{\textcolor{blue}{16.2}}$           & ~0.3$_{\textcolor{blue}{0.4}}$            & ~0.0$_{\textcolor{blue}{0.0}}$            & ~0.1$_{\textcolor{blue}{0.2}}$      & 19.4      \\
                       & Fisher        & ~0.0$_{\textcolor{blue}{0.0}}$            & 12.9$_{\textcolor{blue}{20.0}}$          & ~1.0$_{\textcolor{blue}{1.7}}$            & 16.9$_{\textcolor{blue}{18.3}}$           & ~0.3$_{\textcolor{blue}{0.4}}$            & ~0.1$_{\textcolor{blue}{0.2}}$            & ~0.2$_{\textcolor{blue}{0.3}}$       & 20.1     \\
                       & LIP           & ~0.0$_{\textcolor{blue}{0.0}}$            & 13.8$_{\textcolor{blue}{21.4}}$          & ~1.6$_{\textcolor{blue}{2.8}}$            & 15.7$_{\textcolor{blue}{17.0}}$           & ~0.2$_{\textcolor{blue}{0.3}}$            & ~0.0$_{\textcolor{blue}{0.0}}$            & ~0.3$_{\textcolor{blue}{0.5}}$       & 20.3     \\
                       & EMMN          & ~0.0$_{\textcolor{blue}{0.0}}$            & 81.2$_{\textcolor{blue}{100.0}}$            & ~1.2$_{\textcolor{blue}{2.1}}$            & 56.7$_{\textcolor{blue}{61.4}}$           & ~0.3$_{\textcolor{blue}{0.4}}$            & ~0.1$_{\textcolor{blue}{0.2}}$            & ~0.2$_{\textcolor{blue}{0.3}}$      & 37.8      \\
                       & CLIP-LIP          &      7.4$_{\textcolor{blue}{10.7}}$     &   35.5$_{\textcolor{blue}{55.1}}$          &  10.4$_{\textcolor{blue}{18.2}}$           &   63.1$_{\textcolor{blue}{68.3}}$        &      30.4$_{\textcolor{blue}{43.9}}$       &      20.6$_{\textcolor{blue}{39.7}}$      &  19.8$_{\textcolor{blue}{31.8}}$      & 49.5      \\
                       & \textbf{Ours} & \textbf{~8.8}$_{\textcolor{blue}{12.7}}$  & \textbf{63.2}$_{\textcolor{blue}{98.1}}$  & \textbf{48.8}$_{\textcolor{blue}{85.2}}$  & \textbf{87.7}$_{\textcolor{blue}{94.9}}$  & \textbf{58.1}$_{\textcolor{blue}{84.0}}$  & \textbf{45.7}$_{\textcolor{blue}{88.1}}$  & \textbf{41.2}$_{\textcolor{blue}{66.1}}$ & \textbf{86.2} \\\bottomrule
\end{tabular}}
\end{subtable}
\begin{subtable}{0.48\textwidth}
\centering
\resizebox{\linewidth}{!}{
\begin{tabular}{c|c|ccccccc|c}
\toprule
\multirow{2}{*}{Backbone} &
  \multirow{2}{*}{Method} &
  \multicolumn{2}{c}{CIFAR} &
  \multirow{2}{*}{Food} &
  \multirow{2}{*}{STL} &
  \multirow{2}{*}{ObjectNet} &
  \multicolumn{2}{c}{ImageNet} &
  \multirow{2}{*}{Score}\\
                       &               & Target  ($\downarrow$)       & Retain ($\uparrow$)        &               &               &               & Target        & All &         \\\hline
\multirow{7}{*}{RN50}  & \textcolor{mygray}{Original}      & \textcolor{mygray}{54.0}         & \textcolor{mygray}{73.4}          & \textcolor{mygray}{60.9}          & \textcolor{mygray}{92.0}          & \textcolor{mygray}{68.9}          & \textcolor{mygray}{52.3}          & \textcolor{mygray}{59.8} & --          \\
                       & FT            & ~0.0$_{\textcolor{blue}{0.0}}$           & 56.3$_{\textcolor{blue}{76.7}}$          & ~1.3$_{\textcolor{blue}{2.1}}$            & 29.3$_{\textcolor{blue}{31.8}}$           & ~0.4$_{\textcolor{blue}{0.6}}$            & ~0.0$_{\textcolor{blue}{0.0}}$            & ~0.2$_{\textcolor{blue}{0.3}}$ &      30.2       \\
                       & GA            & ~0.0$_{\textcolor{blue}{0.0}}$           & 20.2$_{\textcolor{blue}{27.5}}$           & ~1.2$_{\textcolor{blue}{2.0}}$            & 19.3$_{\textcolor{blue}{21.0}}$           & ~0.3$_{\textcolor{blue}{0.4}}$            & ~0.0$_{\textcolor{blue}{0.0}}$            & ~0.1$_{\textcolor{blue}{0.2}}$         & 21.6  \\
                       & Fisher        & ~0.0$_{\textcolor{blue}{0.0}}$            & 12.4$_{\textcolor{blue}{16.9}}$           & ~1.2$_{\textcolor{blue}{2.0}}$            & ~13.9$_{\textcolor{blue}{21.0}}$           & ~0.2$_{\textcolor{blue}{0.4}}$            & ~0.0$_{\textcolor{blue}{0.0}}$            & ~0.2$_{\textcolor{blue}{0.2}}$        & 19.2    \\
                       & LIP           & ~0.0$_{\textcolor{blue}{0.0}}$            & 13.9$_{\textcolor{blue}{18.9}}$           & ~0.9$_{\textcolor{blue}{1.5}}$            & 12.3$_{\textcolor{blue}{13.4}}$           & ~0.4$_{\textcolor{blue}{0.6}}$            & ~0.1$_{\textcolor{blue}{0.2}}$            & ~0.2$_{\textcolor{blue}{0.3}}$        & 19.3    \\
                       & EMMN          & ~0.0$_{\textcolor{blue}{0.0}}$            & 48.5$_{\textcolor{blue}{66.1}}$           & ~1.3$_{\textcolor{blue}{2.1}}$            & 21.2$_{\textcolor{blue}{23.0}}$           & ~0.2$_{\textcolor{blue}{0.3}}$            & ~0.0$_{\textcolor{blue}{0.0}}$            & ~0.1$_{\textcolor{blue}{0.2}}$       & 27.4     \\
                       & CLIP-LIP          &   3.8$_{\textcolor{blue}{7.0}}$        &   45.3$_{\textcolor{blue}{61.7}}$         &   16.6$_{\textcolor{blue}{27.3}}$          &   74.2$_{\textcolor{blue}{80.7}}$        &     33.6$_{\textcolor{blue}{48.8}}$        &       28.9$_{\textcolor{blue}{55.3}}$     &      39.5$_{\textcolor{blue}{66.1}}$      & 61.8  \\
                       & \textbf{Ours} & ~\textbf{0.9}$_{\textcolor{blue}{1.7}}$  & \textbf{63.2}$_{\textcolor{blue}{86.1}}$  & \textbf{50.5}$_{\textcolor{blue}{82.9}}$  & \textbf{84.4}$_{\textcolor{blue}{91.7}}$  & \textbf{56.4}$_{\textcolor{blue}{81.9}}$  & \textbf{51.5}$_{\textcolor{blue}{98.5}}$  & \textbf{52.1}$_{\textcolor{blue}{87.1}}$  & \textbf{89.5} \\\hline
\multirow{7}{*}{RN101} & \textcolor{mygray}{Original}      & \textcolor{mygray}{91.4}         & \textcolor{mygray}{79.6}          & \textcolor{mygray}{57.3}          & \textcolor{mygray}{92.4}          & \textcolor{mygray}{69.2}          & \textcolor{mygray}{50.6}          & \textcolor{mygray}{62.3} & --       \\
& FT            & ~0.0$_{\textcolor{blue}{0.0}}$          & 58.3$_{\textcolor{blue}{73.2}}$          & ~1.4$_{\textcolor{blue}{2.4}}$           & 40.4$_{\textcolor{blue}{43.7}}$          & ~0.3$_{\textcolor{blue}{0.4}}$           & ~0.1$_{\textcolor{blue}{0.2}}$           & ~0.3$_{\textcolor{blue}{0.5}}$        & 31.5   \\
& GA            &  ~0.0$_{\textcolor{blue}{0.0}}$         & 20.2$_{\textcolor{blue}{25.4}}$           & ~1.2$_{\textcolor{blue}{2.1}}$          & 19.3$_{\textcolor{blue}{20.9}}$          & ~0.2$_{\textcolor{blue}{0.3}}$           & ~0.0$_{\textcolor{blue}{0.0}}$           &    ~0.1$_{\textcolor{blue}{0.2}}$      & 21.3  \\
& Fisher        & ~0.0$_{\textcolor{blue}{0.0}}$           & 13.1$_{\textcolor{blue}{16.5}}$          & ~1.0$_{\textcolor{blue}{1.7}}$           & 14.2$_{\textcolor{blue}{15.4}}$          & ~0.3$_{\textcolor{blue}{0.4}}$           & ~0.1$_{\textcolor{blue}{0.2}}$           & ~0.1$_{\textcolor{blue}{0.2}}$       & 19.2   \\
& LIP           & ~0.0$_{\textcolor{blue}{0.0}}$           & 15.7$_{\textcolor{blue}{19.7}}$         & ~1.6$_{\textcolor{blue}{2.8}}$           & 15.7$_{\textcolor{blue}{17.0}}$          & ~0.3$_{\textcolor{blue}{0.4}}$           & ~0.0$_{\textcolor{blue}{0.0}}$           & ~0.2$_{\textcolor{blue}{0.3}}$      & 20.0     \\
& EMMN          & ~0.0$_{\textcolor{blue}{0.0}}$          & 48.3$_{\textcolor{blue}{60.7}}$        & ~1.4$_{\textcolor{blue}{2.4}}$           & 35.8$_{\textcolor{blue}{38.7}}$          & ~0.1$_{\textcolor{blue}{0.1}}$           & ~0.1$_{\textcolor{blue}{0.2}}$           & ~0.2$_{\textcolor{blue}{0.3}}$      & 28.9     \\
& CLIP-LIP          &    9.6$_{\textcolor{blue}{10.5}}$      &       52.8$_{\textcolor{blue}{66.3}}$    &    19.5$_{\textcolor{blue}{34.0}}$        &    76.0$_{\textcolor{blue}{82.3}}$      &      50.7$_{\textcolor{blue}{73.3}}$      &      42.5$_{\textcolor{blue}{84.0}}$     &  43.6$_{\textcolor{blue}{70.0}}$        & 71.3   \\
& \textbf{Ours} & ~\textbf{3.2}$_{\textcolor{blue}{3.5}}$ & \textbf{\textcolor{blue}{70.3}}$_{\textcolor{blue}{88.4}}$ & \textbf{46.2}$_{\textcolor{blue}{80.6}}$ & \textbf{86.6}$_{\textcolor{blue}{93.7}}$ & \textbf{58.3}$_{\textcolor{blue}{84.2}}$ & \textbf{43.9}$_{\textcolor{blue}{86.8}}$ & \textbf{46.6}$_{\textcolor{blue}{74.8}}$ & \textbf{86.4}\\\bottomrule
\end{tabular}}
\end{subtable}

\end{table*}
\begin{figure*}
    \centering
    \includegraphics[width=0.9\linewidth]{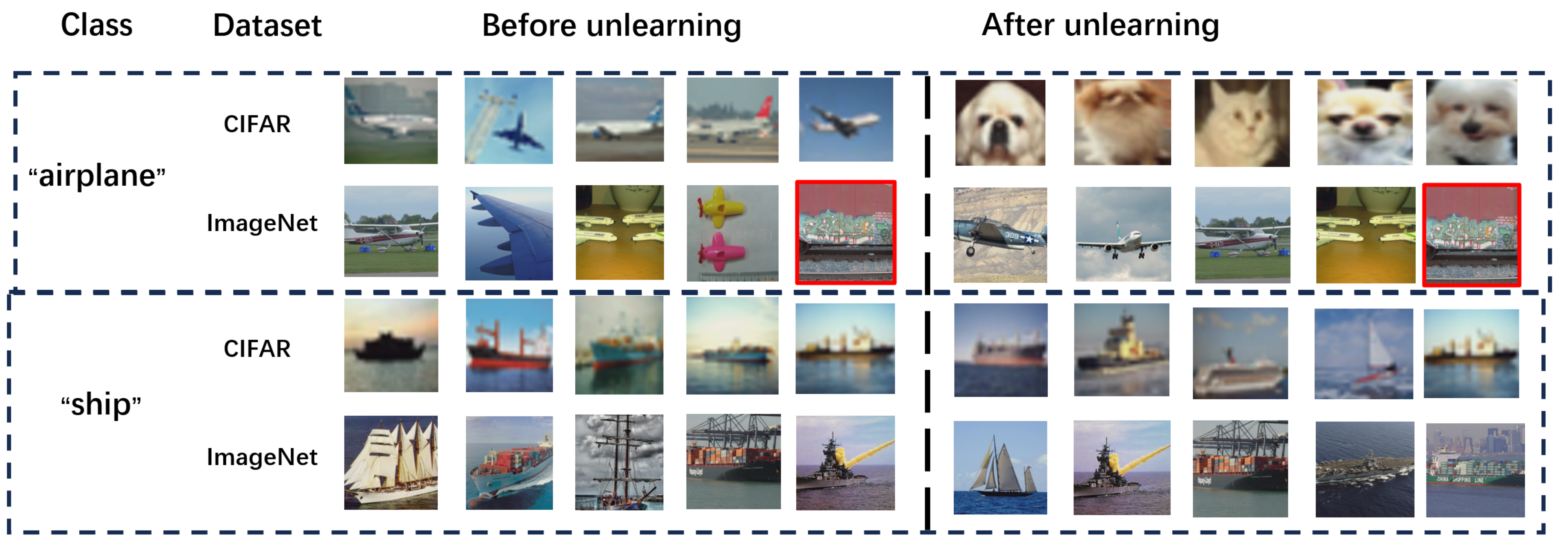}
    \caption{Retrieval results of the CLIP pre- and post-unlearning process.}
    \label{fig:unlearn_airplane}
\end{figure*}

The Results are shown in \cref{tab:breeds_forget}. We observe that both the Fisher and LIP methods fail to optimize the model, resulting in severe catastrophic forgetting across the tested datasets. In comparison, FT, GA, EMMN,and CLIP-LIP perform slightly better but still suffer from catastrophic forgetting to varying degrees. For instance, optimizing only the retaining set with the FT strategy enables the CLIP model to forget the targeted subgroup while retaining knowledge of the intended subgroups. However, this approach leads to noticeable forgetting across other datasets, including different superclasses in ImageNet and unrelated datasets. Additionally, when applying gradient ascent to the forgetting subgroup with the GA  strategy, CLIP exhibits forgetting within other subgroups of the same superclasses, and its zero-shot performance deteriorates due to catastrophic forgetting. Even with the inclusion of both forgetting and retaining samples in the fine-tuning process under the EMMN strategy, these issues remain inadequately addressed. By incorporating layer importance relative to the forgetting dataset to minimize the adverse effects of fine-tuning and leveraging a model merging strategy, we can preserve most of the model's zero-shot performance while ensuring that the targeted subgroup is forgotten.

We also present the retrieval results using our fine-tuned CLIP model, which has forgotten the target subgroup, in ~\cref{fig:unlearn_monkey}. Before unlearning, when using the prompt 'monkey' to retrieve images, the model retrieves images from various subgroups, including 'marmoset' and 'howler monkey,' among others. Additionally, when prompted with specific subclass names, as shown in the second and third rows in figure, the model retrieves corresponding subgroup images. After unlearning, we observe that images of 'marmoset' are no longer retrieved when prompted with either the superclass name 'monkey' or the specific subgroup name 'marmoset,' indicating that the knowledge of 'marmoset' has been effectively erased from CLIP's parameter space. We also include retrieval results when using the prompt 'panda'; the consistent performance between the pre- and post-unlearning versions of CLIP shows that knowledge of other classes remains largely unaffected.

\subsection{Cross-domain forgetting in CIFAR-10}

We then validate the effectiveness of our method in forgetting images belonging to common classes in CIFAR-10, whose concepts also appear in other datasets, such as ImageNet. The main challenge lies in the significant distributional differences between CIFAR-10 and the pre-trained dataset, which can result in severe catastrophic forgetting during the fine-tuning process for unlearning. In this experiment, we select 'airplane' and 'truck' as the target groups to unlearn from CLIP. In addition to evaluating the zero-shot classification performance of these targeted classes in CIFAR-10, we also assess their performance in ImageNet to determine whether unlearning is accurately confined to the specified distribution, without affecting the entire concept.

We present the results in \cref{tab:forget_cifar}. Compared to the results on ImageNet, models fine-tuned on the CIFAR-10 dataset exhibit more severe catastrophic forgetting, even with methods like FT, GA, and EMMN, which retain some zero-shot capability after forgetting specific classes in ImageNet. Notably, models fine-tuned on CIFAR-10 to retain performance demonstrate strong zero-shot performance on STL-10, a dataset with a distribution similar to CIFAR-10, highlighting the impact of the fine-tuning dataset on model performance. In contrast, our method shows remarkable robustness under such a significant distribution shift. Specifically, while the post-unlearning CLIP achieves random classification performance on the forgotten class, it retains at least $66.1\%$ of its original zero-shot performance.

We also retrieve related classes in CIFAR-10 and ImageNet using the pre- and post-unlearning CLIP models for the 'airplane' class. As shown in \cref{fig:unlearn_airplane}, we observe that the post-unlearning CLIP is unable to retrieve airplanes from the CIFAR-10 dataset but can still retrieve airplanes from the ImageNet dataset, indicating precise forgetting limited to the target distribution. The consistency of retrieved results between the pre- and post-unlearning CLIP further demonstrates the effectiveness of our method in preserving unrelated knowledge within the model.

% imagenet结果的位置和前面保持一致性
\subsection{Style forgetting}
\begin{figure}
    \centering
    \includegraphics[width=\linewidth]{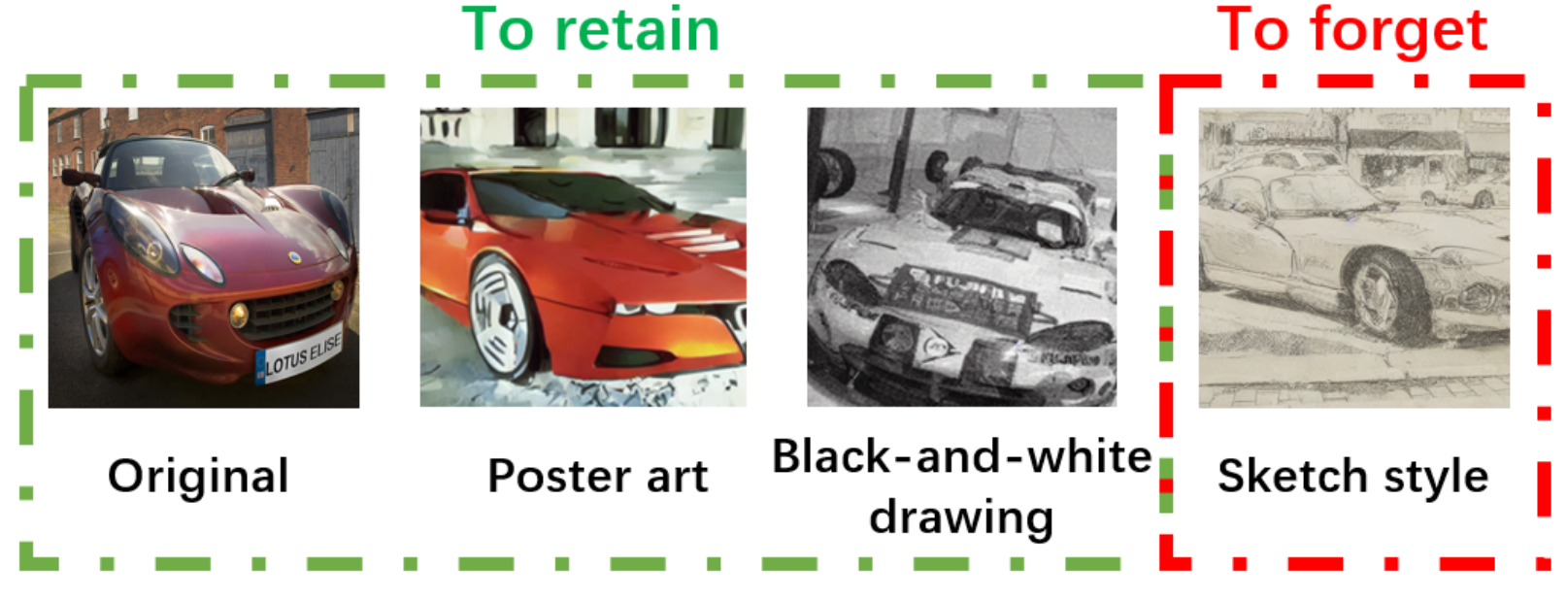}
    \caption{Examples of stylized images of the sports car in the ImageNet-1K dataset.}
    \label{fig:style_car}
\end{figure}

\begin{table}[]
\centering
\caption{Evaluation results of CLIP after forgetting sketch-style "auto-cars" using various methods. }\label{tab:forget_car}
\resizebox{\linewidth}{!}{
\begin{tabular}{c|ccccccc|c}
\toprule
\multirow{2}{*}{Method}  &
  \multirow{2}{*}{CIFAR} &
  \multirow{2}{*}{Food} &
  \multirow{2}{*}{STL} &
  \multirow{2}{*}{ObjectNet} &
  \multicolumn{3}{c}{ImageNet}&
  \multirow{2}{*}{Score}\\
                &                          &           &           &           & Target style  ($\downarrow$)  & Retain  ($\uparrow$)    & All ($\uparrow$)          \\\hline
Original        & 60.8     & 52.2      & 88.4      & 65.4     &  68.1      & 82.5  & 58.6  & --    \\
FT              & 10.1$_{\textcolor{blue}{16.6}}$     &   ~1.9$_{\textcolor{blue}{3.6}}$    &  10.5$_{\textcolor{blue}{11.9}}$     &   3.9$_{\textcolor{blue}{6.0}}$   &   99.5$_{\textcolor{blue}{100.0}}$       & 98.3$_{\textcolor{blue}{100.0}}$  &  ~1.1$_{\textcolor{blue}{1.9}}$   &   20.0  \\
GA             &   46.7$_{\textcolor{blue}{76.8}}$    &  39.8$_{\textcolor{blue}{76.2}}$     &  83.2$_{\textcolor{blue}{94.1}}$     &     57.8$_{\textcolor{blue}{88.4}}$      &   ~2.0$_{\textcolor{blue}{2.9}}$      & 50.3$_{\textcolor{blue}{61.0}}$ &   52.1$_{\textcolor{blue}{88.9}}$    & 83.2  \\
Fisher         &   10.5$_{\textcolor{blue}{17.3}}$    &   ~1.5$_{\textcolor{blue}{2.9}}$      &    15.9$_{\textcolor{blue}{18.0}}$      &  1.1$_{\textcolor{blue}{1.7}}$         &   ~0.1$_{\textcolor{blue}{0.1}}$     & ~0.5$_{\textcolor{blue}{0.6}}$      &  ~0.2$_{\textcolor{blue}{0.3}}$   & 20.1   \\
LIP            &   11.2$_{\textcolor{blue}{18.4}}$    &   ~0.9$_{\textcolor{blue}{1.7}}$       &   13.6$_{\textcolor{blue}{15.4}}$    &   4.2$_{\textcolor{blue}{6.4}}$        &   ~0.0$_{\textcolor{blue}{0.0}}$     & ~0.0$_{\textcolor{blue}{0.0}}$   &   ~0.5$_{\textcolor{blue}{0.9}}$  & 20.4  \\
EMMN           &   16.3$_{\textcolor{blue}{26.8}}$    &  26.5$_{\textcolor{blue}{50.8}}$       &    39.2$_{\textcolor{blue}{44.3}}$       &  20.1$_{\textcolor{blue}{30.7}}$         &     14.5$_{\textcolor{blue}{21.3}}$      & 70.2$_{\textcolor{blue}{85.1}}$     &  ~4.8$_{\textcolor{blue}{8.2}}$ & 46.4    \\
CLIP-LIP            &   15.4$_{\textcolor{blue}{25.3}}$    &   ~6.5$_{\textcolor{blue}{12.5}}$       &   18.7$_{\textcolor{blue}{21.2}}$    &   8.4$_{\textcolor{blue}{12.8}}$        &   1.6$_{\textcolor{blue}{2.3}}$     & 2.4$_{\textcolor{blue}{2.9}}$   &   1.3$_{\textcolor{blue}{2.2}}$ & 24.9    \\
\textbf{Ours} & \textbf{49.2}$_{\textcolor{blue}{80.9}}$ & \textbf{43.0}$_{\textcolor{blue}{82.4}}$ &     \textbf{80.8}$_{\textcolor{blue}{91.4}}$     & \textbf{58.9}$_{\textcolor{blue}{90.1}}$ & \textbf{~2.1}$_{\textcolor{blue}{3.1}}$ & \textbf{66.0}$_{\textcolor{blue}{80.0}}$ & \textbf{51.9}$_{\textcolor{blue}{88.6}}$ & \textbf{87.2}\\\bottomrule
\end{tabular}}
\end{table}

We further evaluate performance on  forgetting a specific style within a given class, representing a more fine-grained subgroup image unlearning challenge. Specifically, we use Stable Diffusion to generate or transform existing images into three distinct styles: poster art, black-and-white drawing, and sketch, as illustrated in Figure~\ref{fig:style_car}. We first fine-tune the CLIP on images with different styles, of which the zero-shot performance is reported at the first line in \cref{tab:forget_car}. Then, different unlearning methods are applied to erase the model's knowledge of cars in the sketch style, while retaining images of cars in the other styles. Results are shown in Table~\ref{tab:forget_car}.

Among the baselines, we find that GA performs the best, effectively forgetting the targeted style while retaining most unrelated knowledge, as indicated by its strong zero-shot performance across different datasets. In comparison, simply fine-tuning CLIP on the retaining set leads to severe catastrophic forgetting, as shown in the results of FT and EMMN. The EMMN results also suggest that incorporating the retaining set into the unlearning process is not always beneficial. Although using the retaining set could potentially reduce model over-forgetting, an imprecise selection of the retaining set and an imbalanced loss between it and the forgetting set can degrade the model's generalization ability on other datasets. By selecting the optimal layer for fine-tuning to minimize the impact on the model's original performance, our method achieves the best overall results, maintaining at least $80.4\%$ of the model's zero-shot capacity.

\subsection{Ablation study and discussion}
\noindent \textbf{On the importance of reminding stage}. The reminding stage follows the forgetting stage, using the retaining set to partially recover any unintended knowledge loss. Notably, our retaining set typically differs from the original pre-training set, potentially further compromising CLIP’s zero-shot classification capability. To examine this, we conduct an ablation study on the impact of iteration count during the reminding stage. Using CIFAR-10 as the unlearning dataset for CLIP, we evaluate performance on both the retain set (ACC$_{r}$) and ImageNet (ACC$_{in}$). As shown in \cref{tab:aba_retain}, we observe that with an increasing number of steps, classification accuracy on the retain set initially improves before declining, while accuracy on the ImageNet-1K dataset consistently decreases, with a significant drop at higher iteration counts. This suggests that while additional steps in the reminding stage help the model relearn knowledge that was unintentionally forgotten, they also lead to a larger distribution shift from the original parameter space, ultimately impairing zero-shot performance.
\begin{table}[]
    \centering
    \caption{Ablation study on the used number of steps at retaining stage on the zero-shot capacity of CLIP. We set the batch size as $64$ for fine-tuning and $\alpha$ as $0.65$ for model merging. }
    \label{tab:aba_retain}
    \resizebox{0.85\columnwidth}{!}{
    \begin{tabular}{c|cccccccc}
    \toprule
        \# steps  & 5 & 10  & 20 & 30 & 40 & 50 & 100  \\\hline
        Acc$_{r}$ & 59.8 &  63.2 & 65.6 & 41.4 & 35.4& 34.9 & 32.5\\
        Acc$_{in}$&  44.9 & 41.2 & 38.1 & 25.8 & 8.5 & ~2.1 & ~0.8\\\bottomrule
    \end{tabular}}
\end{table}

\noindent \textbf{On the model merging}. In our work, we use model merging to fully restore zero-shot capacity. While our method leverages a calibration dataset to identify the optimal coefficient, $\alpha$, we also conduct an additional experiment by directly varying the coefficients to evaluate both forgetting performance (ACC$_{f}$) and recovery performance (ACC$_{r}$/ACC$_{in}$). As shown in \cref{tab:aba_restore}, achieving an optimal $\alpha$ is crucial to balancing forgetting and retaining performance. With increasing $\alpha$, the model’s retaining performance and zero-shot classification capacity improve continuously. However, a large $\alpha$ can also cause the model to relearn data that should be forgotten, as indicated by the $68.8\%$ accuracy on the forgetting set when $\alpha$ is set to $0.9$.

\begin{table}[]
    \centering
    \caption{Ablation study on the $\alpha$ for the model merging operation at the restoring stage to the zero-shot capacity of CLIP. }
    \label{tab:aba_restore}
    \resizebox{0.75\columnwidth}{!}{
    \begin{tabular}{c|cccccc}
    \toprule
        $\alpha$ & 0.1  & 0.3  & 0.5 & 0.65& 0.7   & 0.9  \\\hline
        Acc$_{f}$ & ~0.0& ~0.0 & ~2.4& ~8.8&  12.8 & 68.8     \\
        Acc$_{r}$ & 22.7& 27.5 & 42.7& 63.2&  66.7 & 78.7     \\
        Acc$_{in}$& ~1.0& ~4.4 & 22.39& 41.2& 43.5 & 60.5     \\\bottomrule
    \end{tabular}}
\end{table}

\section{Conclusion}
In this paper, we study the problem of forgetting a subgroup of images in the CLIP model. We propose a three-stage method consisting of the forgetting, retaining, and restoring steps. Our method does not require any pre-trained dataset, allowing it to achieve a balanced approach by selectively forgetting the targeted subgroup of images while preserving the knowledge that should not be forgotten. Experimental results validate the effectiveness of our method and highlight the challenges of this task, as most classical methods fail.

\noindent \textbf{Acknowledgment}. This project is supported by the gift funding from Cisco Research.

{
    \small
    \bibliographystyle{ieeenat_fullname}
    \bibliography{main}
}


\clearpage
\appendix
\renewcommand{\appendixpagename}{Appendix}
\appendixpage

\section{Experiment details}
\label{sec:exp_de}

\subsection{Implementations}
\label{sec:imp_base}

In our work, we implement the baseline methods under the setting of CLIP models, which are initially designed for conventional machine unlearning methods. Here, we present the implementation details as follows.

\noindent \textbf{Notations}. Given the CLIP model denoted as \( g(\cdot, \cdot) = \{g^{\text{img}}(\cdot), g^{\text{txt}}(\cdot)\} \), processes image and text inputs into embeddings, \( e^{\text{img}} \) and \( e^{\text{txt}} \). We denote the forgetting dataset as $\mathcal{D}_{f}$, the retaining dataset as $\mathcal{D}_{r}$. Each example $x^{img}$ in both $\mathcal{D}_{f}$ and $\mathcal{D}_{r}$ has the label $c$, which is used as the paired text description for text encoding in CLIP.  For zero-shot classification, we compute the text embedding of different classes. For each image, we compute its image embedding and the cosine similarity between the image embedding and text embedding. We take the class which has the maximum cosine similarity as the classification result.

\noindent \textbf{FT}.  We directly fine-tune the image encoder $g^{\text{img}}(\cdot)$ on the retaining dataset. We use the Adam optimizer and set the learning rate as $10^{-6}$. The batch size is set as $128$.  We optimize the model on the retaining set with $2$ epochs.

\noindent \textbf{GA}. We perform the gradient ascent on the forgetting dataset, setting the learning rate as $10^{-6}$. We use the Adam optimizer the optimize the model with $2$ epochs.

\noindent \textbf{Fisher}. We compute the Fisher Information Matrix on the forgetting dataset and perturb the model parameters with noise sampled from a Gaussian distribution, where the variance is derived from the Fisher Information Matrix.

\noindent \textbf{LIP}. For a given image to forget,  we first generate its multiple copies with injected random noise. Then, we optimize the embedding of original images and noisy images based on lipschitz constraint as follows,
\begin{equation}
    \min \mathcal{L}_{emb} = \sum_{i=1}^{N}\frac{1}{N}\frac{\Vert g^{\text{img}}(x) - g^{\text{img}}(x+\epsilon_{i}) \vert}{\Vert \epsilon \Vert},
\end{equation}
where we have $\epsilon \sim \mathcal{N}(0, \sigma)$.

At the same time, under the setting of zero-shot classification, we also apply the lipschitz constraint on the cosine similarity between the image embedding and text embedding as follows,
\begin{equation}
    \min \mathcal{L}_{cls} = \sum_{i=1}^{N} \frac{1}{N}\frac{\Vert l(x) - l(x+\epsilon) \vert}{\Vert \epsilon \Vert},
\end{equation}
where $l(\cdot)$ is the cosine similarity between the image embedding and the text embedding computed by possible classes.

\noindent \textbf{EMMN}. We jointly optimize $g^{\text{img}}(\cdot)$ on the forgetting dataset and the retaining dataset. Specifically, we maximize the loss function value on the forgetting dataset and minimize the loss function value on the retaining dataset. The learning rate is set to $1 \times 10^{-6}$. We use the Adam optimizer and fine-tune the model for $5$ epochs.

\subsection{Model merging for continuous forgetting}
\label{sec:conti}

In our method, we leverage model merging to restore the zero-shot classification capacity of the studied model while forgetting the targeted knowledge. Through our experimental exploration, we observe that two models, each forgetting one class, can be simply merged into a single model capable of forgetting both classes.We report the result in \cref{tab:my_label}. 

First, we use our method to obtain CLIP models that have forgotten the classes "ship," "airplane," and "cat," respectively. Next, we merge the models forgetting "ship" and "airplane" (denoted as S+A). The resulting model successfully forgets both classes. Similarly, the model denoted as S+A+C, obtained by merging models that forget "ship," "airplane," and "cat," successfully forgets all three classes together. An intriguing observation is that while the model performs worse on most classes, it exhibits improved performance on certain specific classes, such as \textit{frog}. Understanding the underlying reasons for this phenomenon remains an open question for future research.

\begin{table}[]
    \centering
    \caption{Leveraging the model merging to achieve continuous unlearning for multiple classes.}
    \label{tab:my_label}
    \resizebox{\columnwidth}{!}{
    \begin{tabular}{c|cccccccccc}
    \toprule
        Classes &  Airplane & Mobile & Bird & Cat & Deer & Dog & Frog & Horse & Ship & Truck \\\hline
        Original &  \textcolor{mygray}{81.5} & \textcolor{mygray}{97.6} & \textcolor{mygray}{85.6} & \textcolor{mygray}{68.6} & \textcolor{mygray}{59.4} & \textcolor{mygray}{67.7} & \textcolor{mygray}{48.4} & \textcolor{mygray}{84.3} & \textcolor{mygray}{68.1} & \textcolor{mygray}{54.5}\\\hdashline
        Ship (S) & 78.5 & 82.2 & 85.4 & 59.9 & 41.3 & 62.9 & 47.9 & 73.7 & \textcolor{red}{0.0} & 78.6\\
        Airplane (A) &  \textcolor{red}{0.0} & 87.6 & 78.7 & 72.3 & 48.9 & 51.5 & 46.2 & 78.5 & 60.5 & 64.1  \\ 
        Cat (C)  & 72.7 & 85.6 & 88.6 & \textcolor{red}{0.0} & 42.3 & 56.6 & 47.7 & 75.9 & 51.5 & 44.6 \\\hdashline
        S+A & \textcolor{red}{10.0} & 88.3 &83.0   & 58.4 & 47.5  & 62.5 & 55.4 & 77.9  & \textcolor{red}{10.0}  & 70.8\\
        S+A+C & \textcolor{red}{5.7} & 91.4 & 88.4 & \textcolor{red}{1.9} & 42.8 & 51.5 & 55.5 & 80.1 & \textcolor{red}{5.1} & 55.0\\\bottomrule
    \end{tabular}}
\end{table}

\end{document}